# Photosensor Oculography: Survey and Parametric Analysis of Designs using Model-Based Simulation

Ioannis Rigas, Hayes Raffle, and Oleg V. Komogortsev

*Abstract*—This paper presents a renewed overview of photosensor oculography (PSOG), an eye-tracking technique based on the principle of using simple photosensors to measure the amount of reflected (usually infrared) light when the eye rotates. Photosensor oculography can provide measurements with high precision, low latency and reduced power consumption, and thus it appears as an attractive option for performing eye-tracking in the emerging head-mounted interaction devices, e.g. augmented and virtual reality (AR/VR) headsets. In our current work we employ an adjustable simulation framework as a common basis for performing an exploratory study of the eye-tracking behavior of different photosensor oculography designs. With the performed experiments we explore the effects from the variation of some basic parameters of the designs on the resulting accuracy and cross-talk, which are crucial characteristics for the seamless operation of human-computer interaction applications based on eye-tracking. Our experimental results reveal the design trade-offs that need to be adopted to tackle the competing conditions that lead to optimum performance of different eye-tracking characteristics. We also present the transformations that arise in the eye-tracking output when sensor shifts occur, and assess the resulting degradation in accuracy for different combinations of eye movements and sensor shifts.

I. INTRODUCTION

THE recording of human eye movements provides a bi-directional source of information related to the *perception* and *intention* of a person. In the past, eye movements have been examined in relation to the mechanisms of visual attention [1, 2], and while exploring the perceptual procedures that support high-level cognitive functions [3, 4]. Furthermore, the recording of a person's gaze and the related eye movement-dynamics have been proposed as alternative means for building input interfaces in the domain of human-computer

I. Rigas and O.V. Komogortsev are with Texas State University, Department of Computer Science, 601 University Dr, San Marcos, TX 78666 USA, (e-mails: rigas.ioann@gmail.com; ok@txstate.edu).
H. Raffle is with Google, 1600 Amphitheater Drive, Mountain View, CA 94043, USA, (e-mail: hraffle@google.com).

interaction (HCI) [5, 6]. More specifically, the integration of eye-tracking functionality in some of the emerging human-computer interaction devices (e.g. AR/VR headsets) seems to be an essential necessity for enabling eye movement-based applications, like foveated rendering [7] and saccade contingent updating [8]. Given the current trends for miniaturization of these devices, the examination of simple and efficient eye-tracking techniques is critical in the effort to advance the developments in this technological field.

The history of eye-tracking systems dates back to the late 1800s (a thorough review can be found in [9]). Early eye-tracking devices were quite invasive. For example, the device described by psychologists E.B. Delabarre [10] and E.B. Huey [11] was based on an eye-attachment with an embedded lever that could mechanically record eye movements on a rotating smoked drum. Delabarre also described the possibility to attach a mirror (instead of lever) that will reflect light and then record the reflected light on a photographic plate, however, practical issues prevented him from using this approach.

In 1901, Dodge and Cline [12] described the creation of a non-invasive eye-tracking device based on a camera. The system was based on the observation that when light hits the eye cornea, a distinct reflection image is generated that can be recorded using a photographic camera and used to monitor eye movements. This operating principle can be considered the predecessor of many of the current video oculography (VOG) techniques. The general approach followed in most modern video oculography techniques is based on the use of image processing algorithms for tracking the relative changes in the positions of pupil center and corneal reflection, and thus estimate changes in gaze [13]. The advances in video sensor technology and the increased computational power of modern CPUs have led to the widespread adoption of video oculography, especially for building remote capturing setups in research lab environments. Video oculography can provide good accuracy during gaze estimation (about 0.5°) and relative robustness to small head movements, achieved via suitable algorithmic processing of the relative positions of the tracked eye features.

A different technique for recording eye movements is based on the measurement of differences in the electric potential between the cornea and the retina [14] when the eye moves. This technique is known as electro-oculography (EOG) and it requires the placement of a few electrodes in the areas around the eye to record the electric signal caused by the eye rotations. A characteristic of this technique is that it can capture eye movements even with the eyes closed. Some disadvantages of the EOG technique is the reduced precision (sensitivity) in levels of 1°-2°, the signal artifacts caused by extraneous bioelectric sources, and the need for frequent measurement of skin impedance since it can affect the recorded values.

Another technique for measuring eye movements is based on the use of a contact lens with an embedded coil moving in a magnetic field [15]. The great advantage of this technique is the very good precision that can be achieved (levels of 0.05°). However, the scleral coil technique is based on an invasive eye-attachment and the use for an extended period (>30 min.) can be discomforting due to the requirement of topical anesthesia when using the contact lens. Also, the external coils that create the necessary magnetic field are bulky, limiting the use of the scleral coil technique in controlled environments.

Our current work explores and analyzes the characteristics of the technique of photosensor oculography (PSOG) [16, 17]. Photosensor oculography has some distinct advantages over the previously described methods. The technique is based on the use of just a few photosensors (usually infrared) to measure the overall reflected amount of light from selected eye regions. Due to the minimalistic setup and use of simple sensors, the technique supports high sampling rates and has low processing complexity. Thus it can provide low eye-tracking delay and reduced power consumption. These properties are particularly advantageous when compared to video oculography, where the high-speed operation requires the application of sophisticated image processing algorithms on thousands of frames (millions of pixel values). Furthermore, in photosensor oculography the sensors need to be positioned in close proximity to the eye, making the technique by design tailored to head-mounted setups. In contrast to electro-oculography and magnetic scleral coil technique, the recording of eye movements in photosensor oculography is performed in a touchless unobtrusive manner. The precision of the technique is limited mostly by the noise in electronics and has been measured to be less than 0.01° [18]. Expectedly, the employed minimalistic setup comes with some disadvantages. The most notable is the intolerance to sensor shifts relative to the head. This might be assumed to be less of a problem for the case of a head-mounted setup, however, small sensor shifts (slippage) can still occur due to head, face and body movements. A second disadvantage is that due to the eye shape, capturing of signal for the vertical component of eye movement can be more challenging.

The advent of new computing devices for human-computer interaction brings new requirements for eye-tracking including the necessity for higher sampling rate, better precision, lower cost, and reduced power consumption. Such requirements are challenging for video-based eye-tracking systems. Our motivation for the current research is to provide an insight into alternative techniques that can be used either by-themselves or be complementary to video-based oculography systems to meet the needs of emerging computing platforms. Our work focuses on the investigation of photosensor oculograghy and for the first time we perform a parametric analysis of basic photosensor oculography designs using model-based simulation. The exact contribution of our work can be summarized as follows:

- We present a survey of the most important photosensor oculography designs and discuss their basic properties.
- We perform experiments by changing the parameters of photosensor oculography designs and exploring the effects on eye-tracking accuracy and crosstalk.
- We perform experiments and examine the degradation in accuracy when sensor shifts occur.

The performed investigation provides valuable insights for the design factors that can affect the quality of the eye-tracking signal, and aspires to offer useful data that can be used for the engineering of improved photosensor oculography systems.

II. REVIEW OF PHOTOSENSOR OCULOGRAPHY DESIGNS

In this section we present a review of the most significant works related to the technology of photosensor oculography. It should be clarified that the term photosensor oculography (PSOG) is used in this paper in order to describe eye-tracking systems that are based on the principle of *using simple sensors in order to capture the overall amount of reflected light from selected regions of the eye surface*. In some previous works, alternative terms have been used to refer to systems based on this principle, e.g. 'photoelectric technique', 'limbus reflection technique', 'infrared oculography (IROG)', and 'infrared reflectance oculography'. We opted to use the term 'photosensor oculography' as more descriptive of the common operation principle shared by different instantiations of the technique. In Fig. 1, we portray the operation principle of photosensor oculography.

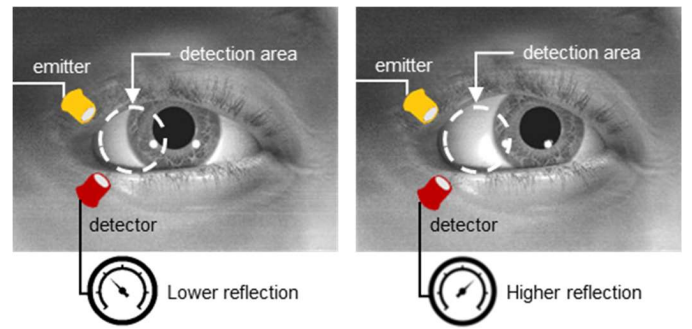

Fig. 1. Graphical presentation of the operation principle of photosensor oculography (PSOG).

Some previous studies overviewing various characteristics of photosensor oculography systems can be found in [17, 19]. Our current survey additionally covers some more recent advances and focuses on the analysis of 'PSOG designs' a term used to denote the detection patterns (areas) used by different instantiations of the technique. The detection patterns and their combination rules are among the most important characteristics of photosensor oculography systems, since they largely determine the eye-tracking properties in terms of linearity, accuracy, cross-talk, and tolerance to shifts. It should be mentioned that in a practical system, the same detection areas can be formed in different ways, for example by using narrow emission and wide detection sensors or vice versa, by using light masking during emission (optical slits) or detection (lens), and by using light guidance via suitable means (e.g. optical fibers).

The basic principle of photosensor oculography was introduced in 1951, in the paper of Torok et al. [16]. The proposed system was a simple and easy-to-use clinical device for inspecting nystagmus, a special type of involuntary rhythmic eye movement. The original setup consisted of a focused beam patterned to form a rectangular capturing area ($PS_1$, Fig. 2) of about 1 mm x 10 mm and positioned to cross the borderline of the sclera (high reflection) and the iris (low reflection), known as the limbus. This setup was suitable for capturing horizontal movements towards the nasal or temporal direction (depending on positioning). To capture vertical movements, as shown in [20], a smaller rectangle ($PS_2$, Fig. 2) of approximate dimensions of 3.7 mm x 1 mm could be rotated to target the border under iris. To avoid disruption of the user the authors proposed the use of infrared (IR) light, a selection adopted in the majority of the following PSOG systems.

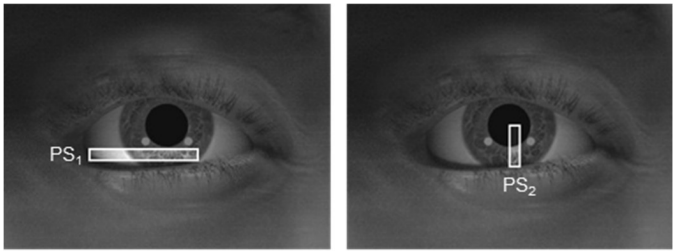

Fig. 2. Detection areas of PSOG designs presented in [16, 20].

The design proposed by Richter [21] some years later, was based on the symmetric positioning of sensor pairs on both sides of the iris. When the eye moves horizontally, the output from one sensor increases while the output of the other decreases. Thus the horizontal output can be calculated by subtracting the outputs from the sensors. When the eye moves vertically, the outputs of sensor pairs change in same way. In this case, the vertical output can be calculated by adding the output from the sensors. The specific design is more robust on the elimination of horizontal-vertical movement cross-talk (interference) than simply taking the relative changes of the output from a single sensor. Furthermore, the design allows capturing larger range of horizontal eye movements on both nasal and temporal direction. It should be noticed that shortly after the initial works of Torok et al. (1951) and Richter (1956), the papers of Smith and Warter [22], and Rassbash [23] (1960), seem to propose (re-introduce) methods based on the principle of photosensor oculography.

A classic system based on the sensor-pair differential design was presented by Stark et al. [24] for the case of capturing horizontal movements. This PSOG design is depicted in Fig. 3 (left). The detection areas seem to cover circular areas with diameter of about 5.2 mm on the both sides of the iris-sclera border. The output can be calculated as $PS_1 - PS_2$. This design allowed capturing horizontal eye movements is the range of ±12°. A variation of this design for capturing simultaneously horizontal and vertical eye movements was described later by Bahill and Stark [25], and the approximate detection areas are presented in Fig 3 (right). The horizontal output can be calculated as $PS_1 - PS_2$ (upper sensors) whereas the vertical output as $PS_3 + PS_4$ (lower sensors). We can observe that the detection areas in this case are narrower (diameter of about 3 mm) in an effort to minimize contact with the eyelids.

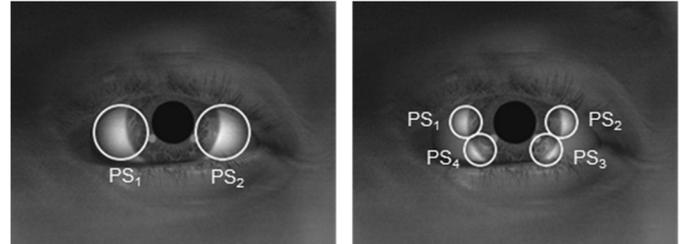

Fig. 3. Detection areas of PSOG designs presented in [24] (left) and [25] (right).

Even with the use of infrared light the PSOG techniques can suffer from ambient light interferences, especially when operated in outdoor environments. For this reason, Wheeless et al. [26] introduced the use of modulated (chopped) light for the emitters of a photosensor system. The detectors were properly configured to perform the respective de-modulation. The detection areas were formed by guiding the light with bundles of optic fiber. The system followed the design that is shown in Fig.4 (left), in this case the detection areas are approximately 1 mm x 3 mm. Capturing of horizontal eye movements was done by subtracting the outputs targeting the sclera-iris boundary ($PS_1 - PS_2$), and capturing of vertical movements was done by subtracting the outputs targeting in the iris-pupil boundary ($PS_3 - PS_4$). The use of differential operation also for the vertical output was opted with the aim to provide robustness for pupil variations. The allowed eye movement ranges were ±15° horizontally and ±10° vertically.

A novel design was proposed by Jones [18], where the two detection areas form a diagonal pattern. In order to form this pattern broad illumination was used, in combination with lenses to de-magnify areas of 1 mm x 7.8 mm on the eye. The method of subtraction-addition of the same sensors was used here to simultaneously capture horizontal and vertical eye movements (sensor measurement 'de-matrixing'). In Fig. 4 (right) we show the detection patterns for this design, the horizontal component is extracted as $PS_1 - PS_2$ and the vertical as $PS_1 + PS_2$. The symmetrical positioning of the detection areas is very important for the accurate operation of this design, and the sensor tilt can be optimized. The shown design corresponds to an angle of 30° from the horizontal line, and the distance between the rectangles' centers is about 8 mm. A modification of the diagonal design was presented in [27], where the size of detection areas was changed to 2.6 mm x 4.6 mm accompanied by other practical improvements such as the use of slits to better control the size and alignment of detection areas, and the use of a calibration procedure to accommodate for asymmetries and different sensor sensitivities. The allowed eye movement ranges for this design were ±18° horizontal and ±9° vertical, and the cross-talk values were measured to be 0.2% and 3.3% respectively.

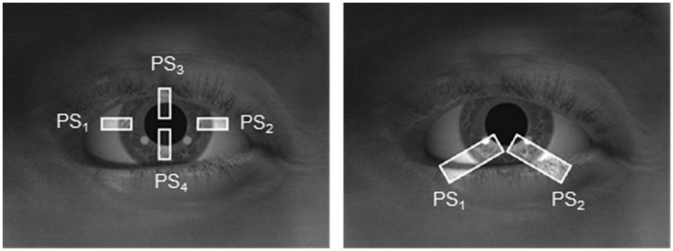

Fig. 4. Detection areas of PSOG designs presented in [26] (left) and [18] (right).

An approach using more sensors was presented by Reulen et al. [28]. A horizontally placed sensor array formed a pattern of nine detection areas ($PS_1$ to $PS_9$) that could be used to capture horizontal eye movements by subtracting the aggregated output of the external sensors ($PS_1 + PS_2$) – ($PS_8 + PS_9$) in order to maximize the captured range. The sensor array could be rotated to capture vertical eye movements. In this case, the output was calculated by subtracting the aggregated outputs of the internal sensors ($PS_2 + PS_3$) – ($PS_7 + PS_8$) to avoid eyelid occlusion. The authors do not clarify the use of other sensors in this paper. The allowed eye movement ranges for this design were ±30° horizontal and ±25°, and the cross-talk was measured to be about 10% in each direction. In Fig. 5 we show a diagram of the potential circular overlapping detection areas of such an array design, with diameters for the areas of about 4 mm. It should be noticed that a variation of the array design was presented in [29] by using a setup for capturing simultaneously horizontal and vertical eye movements. The setup consisted of ten sensor sets (emitter-receiver) for horizontal and four sensor sets for vertical capturing, fitted in a common frame.

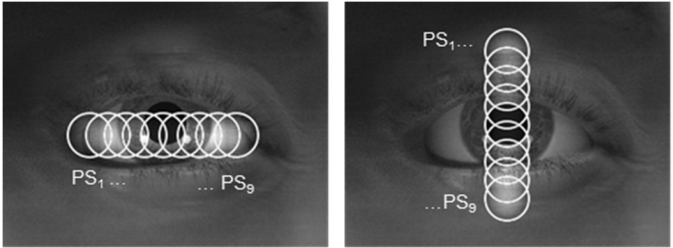

Fig. 5. Detection areas of PSOG array design presented in [28].

Although the last years the main focus has mostly moved towards video oculography, there are various research studies that have employed systems based on the principles of photosensor oculography. Devices based on capturing with photosensor pairs were suggested for facilitating the measurement of eye movements and eye blinks during fMRI experiments [30, 31]. Also, a system that seems to be based on photosensor oculography principles was described in [32] and more recently was used to explore eye movements in behavioral studies for dyslexia [33]. In [34] a simple setup based on infrared photosensors was tested for automotive applications, and specifically, for measuring driver's drowsiness.

### III. SIMULATION EXPERIMENTS

The developed framework for the simulation experiments is based on the following steps: a) generation of synthetic eye images via 3-D rendering while eye movements and sensor shifts are performed, b) processing of the rendered images for simulating the detection areas of PSOG designs and for calculating their output, and c) transformation (calibration) of the PSOG output from raw units to degrees of visual angle.

#### A. Generation of Synthetic Eye Images

The generation of synthetic eye images used for the extraction of PSOG designs is performed using the simulation environment developed in [35]. This environment is built with the software package Blender [36] and provides template models for the eye, camera and light sources, along with a programming interface for positioning and rotating the models. The simulated models can be used to build an eye-tracking setup, and the corresponding eye images are generated via 3-D graphics rendering. The setup that we developed for our experiments used an (left) eye model with an eyeball diameter of 24 mm, iris horizontal diameter of 9.5 mm, and corneal refractive index of 1.336. Pupil diameter was originally set to 4 mm, but during the experiments we used signal captured from real eye movements to accurately simulate dilation in range of [3.6 mm, 4.6 mm]. Two light sources were positioned ±1.4 cm horizontally, 1 cm under and 3 cm away from the eye pupil center (distances refer to left eye in neutral position), and a camera model was placed in central position (horizontally and vertically) and 5 cm away from the pupil center. The field of view of the camera was set to 45°, and the resolution of the images was 240 x 320 pixels.

During the experiments we used eye movement vectors (values to rotate the eye model) captured from real eye movements. These real eye movement vectors were used to increase the realism of simulations and they also served as a ground truth for the evaluation of accuracy and crosstalk presented in Section IV. The real eye movement signals were captured using an EyeLink 1000 eye-tracker [37] (vendor reported accuracy 0.5°) at a sampling rate of 1000 Hz (left eye only). Subject's positioning was at a distance of 550 mm from a computer screen (size 297 x 484 mm, resolution 1050 x 1680 pixels) where the visual stimulus was presented (head was restrained using a head-bar with a forehead). The visual stimulus consisted of a white 'jumping' point of light making horizontal and vertical 'jumps' on a black background, and changing its position every 1 second for a total duration of 36 seconds. The amplitude of 'jumps' was increasing from ±2.5° to ±10°, with steps of 2.5°. In Fig. 6, we show the ground truth signals that were used as input for the simulations.

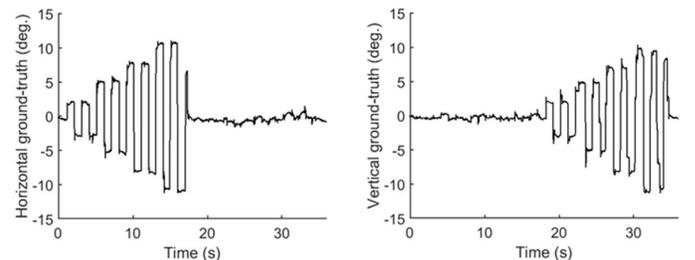

Fig. 6. Real eye movement signals used as ground truth input during the simulation experiments.

## B. Simulation of the Detection Areas and the Output of PSOG Designs

The second step in simulation queue involves the processing of the rendered eye images in order to model the detection areas of PSOG designs and calculate their output.

The output of a single photosensor is simulated by performing a window binning operation (window averaging) on the pixel intensity values of the image region that corresponds to the detection area of the sensor. To simulate simple sensors with circular detection areas (no light masking) the windows are modulated using a Gaussian filter with zero mean and σ = ½ window size. To simulate sensors with rectangular detection areas (light making is assumed) the averaging operation is done directly on the pixel intensity values of the window. The general formula for the calculation of the output of a photosensor is given in Eq. (1) (for non-Gaussian modulated $G_{i,j} \rightarrow 1$). The followed procedure for the simulation of a single photosensor's output is graphically summarized in Fig. 7.

$$I_{PS} = \frac{\sum_{i,j} G_{i,j} \cdot W_{i,j}}{i \cdot j}, \; i,j = pixel\;coord. \quad (1)$$

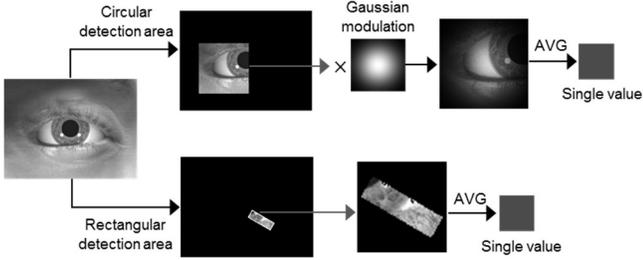

Fig. 7. Presentation of the steps followed to simulate the output of a single photosensor.

The previously described procedure for the simulation of a photosensor's output is supported by the following rationale: in theory, a photosensor can be modeled as a controlled current source connected in parallel to an exponential diode [38]. The currents of the controlled source $I_p$ and the exponential diode $I_d$ can be modeled using Eq. 2-3:

$$I_p = R_\lambda \cdot P \quad (2)$$
$$I_d = I_s \cdot \left( e^{\frac{q \cdot V_A}{k_B \cdot T}} - 1 \right) \quad (3)$$

where $R_\lambda$ is the responsivity at wavelength $\lambda$, $P$ is the incident light power, $I_s$ is the reverse saturation current, $q$ is the electron charge, $V_A$ is the applied bias voltage, $k_B = 1.38 \cdot 10^{-2} \; J/K$ is the Boltzmann constant, and T is the absolute temperature. In photovoltaic mode the photodiode is zero-biased (V = 0), and since $I_d \rightarrow 0$ the output of the sensor is analogous to the incident light power $P$ ($R_\lambda$ can be considered constant for given conditions). The followed simulation procedure assumes that the incident light power from the reflection on a specific detection area of the eye can be resembled by the performed (modulated) window binning operation.

As the basis for our experiments we selected the four most characteristic designs from those described in Section II. For each of these designs (denoted $D_{1-4}$ and shown in Fig. 8) we experimented by changing the values of important parameters, and performed the respective simulations for each parameter set. The values were selected so that we start from a very small detection area and progressively reach dimensions that cover large part of the eye and periocular areas.

For design $D_1$ (Fig. 8, top-left), we change the Length (L) and the Width (W) of the rectangular detection areas $PS_{1-4}$. Both for L and W the chosen ranges were [0.5 mm, 12 mm] with step of 0.5 mm (all values measured on eye surface). The raw output for design $D_1$ is calculated as ($I_{PS1} - I_{PS2}$) for the horizontal output and ($I_{PS3} - I_{PS4}$) for the vertical output, where the values for $I_{PS1-4}$ are calculated using the described window averaging procedure for the new detection areas each time. For design $D_2$ (Fig. 8, top-right), we change the Length (L) and Width (W) of the rectangular detection areas $PS_{1-2}$, and additionally, we change the angle (A) formed by the detection areas with the horizontal plane. For the case of L and W we use values in range [0.5 mm, 12 mm] with step of 0.5 mm, whereas for A the values are in range [5°, 45°] with step of 5°. Design $D_2$ uses the same detection areas for calculating the horizontal and vertical outputs, specifically, the horizontal output is calculated as ($I_{PS1} - I_{PS2}$) and the vertical output as ($I_{PS1} + I_{PS2}$). For design $D_3$ (Fig. 8, bottom-left) the detection areas are circularly modulated, and here the only changing parameter is the diameter D of the detection areas $PS_{1-4}$. The used values for parameter D lie in range [0.5 mm, 12 mm] with a step of 0.5 mm. In this case, the horizontal output is calculated as ($I_{PS1} - I_{PS2}$) and the vertical output as ($I_{PS3} + I_{PS4}$). For design $D_4$ (Fig. 8, bottom-right) the detection areas are formed by horizontal and vertical arrays of sensors. For this reason, except from the diameter (D) of the detection areas we also change the vertical positioning of the horizontal array ($P_y$) and the horizontal positioning of the vertical array ($P_x$). The used values for D are in the range of [0.5 mm, 12 mm] with step of 0.5 mm, and for $P_y$ and $P_x$ in the range of [-2 mm, 2 mm] with step of 0.5 mm (positioning measured from the pupil center in primary position). For design $D_4$ the horizontal output is calculated as ($I_{PS1} + I_{PS2}$) – ($I_{PS8} + I_{PS9}$) with reference to the horizontal array, and the vertical output is calculated as ($I_{PS2} + I_{PS3}$) – ($I_{PS7} + I_{PS8}$) with reference to the vertical array.

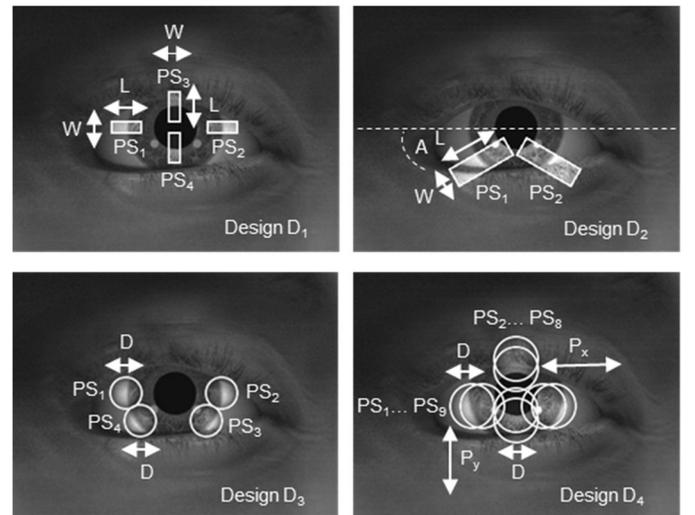

Fig. 8. Designs $D_{1-4}$ used during the experiments and the respective changes made on their parameters.

## C. Calibration Procedure

The raw output values calculated with the previously described operations need to be transformed in degrees of visual angle. This transformation can be performed by modeling a suitable calibration (mapping) function. In our experiments, we model a quadratic calibration function for each component of eye movement separately (horizontal and vertical), as shown in Eq. (4-5):

$$f_C^H(x) = a^H \cdot x^2 + b^H \cdot x + c^H \quad (4)$$
$$f_C^V(y) = a^V \cdot y^2 + b^V \cdot y + c^V \quad (5)$$

To calculate the calibration parameters, we need to use a minimum of three eye positions per direction $(x, y)$, and for this reason we rotate the simulated eye model at the predefined positions of -10°, 0° and 10° horizontally and vertically, and then, perform a Least Squares regression fit on the corresponding raw output values $f_C^H(x), f_C^V(y)$. The estimated calibration parameters are then used to transform any new raw values calculated from the PSOG designs. A separate calibration procedure is performed for each of the designs $D_{1-4}$, and the calibration procedure is repeated every time a change is made on the design parameters.

## IV. RESULTS AND ANALYSIS

### A. The Effects of Design Parameters on Eye-tracking Performance

In this section we present the results from our experiments involving the exploration of the important eye-tracking characteristics of accuracy and crosstalk, when varying the parameters of the PSOG designs.

*Accuracy (Mean Absolute Error)*

Accuracy is assessed by calculating the mean absolute error between the ground truth eye-tracking signals $s_{Gt\,j}^{H,V}$ and the simulated PSOG output signals $s_{Sim\,j}^{H,V}$, as shown in Eq. (6-7). To calculate accuracy, we use only the parts of the signals that correspond to fixations in order to avoid large outliers from saccade transitions.

$$acc_i^H = \sum_{j=1}^{M} \left| s_{Sim\,j}^H - s_{Gt\,j}^H \right| / M, i = 1, \dots, N_H \quad (6)$$
$$acc_i^V = \sum_{j=1}^{M} \left| s_{Sim\,j}^V - s_{Gt\,j}^V \right| / M, i = 1, \dots, N_V \quad (7)$$

In Eq. (6-7), $acc_i^{H,V}$ is the accuracy (horizontal, vertical) for each of the performed fixations in horizontal and vertical direction ($N_{H,V}$), and $M$ denotes the number of samples within each fixation (different for each fixation).

*Crosstalk*

Crosstalk is assessed by calculating the absolute ratio of the observed movement in the signal of one output (channel) when ground truth movements occur on the opposite direction, as shown in Eq. (8-9):

$$cross_i^{HV} = \left| \frac{\sum_{j=1}^{M} s_{Sim\,j}^H - \sum_{j=1}^{M} s_{Gt\,j}^H}{\sum_{j=1}^{M} s_{Gt\,j}^V} \right|, i = 1, \dots, N_V \quad (8)$$

$$cross_i^{VH} = \left| \frac{\sum_{j=1}^{M} s_{Sim\,j}^V - \sum_{j=1}^{M} s_{Gt\,j}^V}{\sum_{j=1}^{M} s_{Gt\,j}^H} \right|, i = 1, \dots, N_H \quad (9)$$

In Eq. (8-9), $cross_i^{HV}$ denotes the crosstalk in horizontal signal when pure vertical movements are performed and $cross_i^{VH}$ is the crosstalk in vertical signal when pure horizontal movements are performed.

In Fig. 9-12 we present the behavior of accuracy and crosstalk when changing the parameters of each design (as explained in Section III). The left column in each figure shows the results for accuracy and the right column shows the results for crosstalk. Since the design parameters can affect jointly the measures of accuracy and crosstalk, the diagrams are in most cases in form of surfaces (except for Fig. 11), where the brighter colors correspond to larger values (worse performance) and the darker colors to smaller values (better performance). In Tables I-II we additionally present quantitative data extracted from the diagrams, which show the optimum performances that can be achieved for each design. We also present the corresponding parameters (dimensions or positioning of detection areas) that lead to these optimum performances. The data presented in Fig. 9-12 and Tables I-II are calculated using Eq. 6-9 and by averaging the values from the respective horizontal or vertical fixations (in Tables I-II we additionally report the standard deviations in parenthesis).

In Fig. 9, we show the diagrams corresponding to design $D_1$ when changing the dimensions of Length (L) and Width (W) of the rectangular detection areas. For the case of accuracy, we can observe that the overall tendency is largely dictated at by Length parameter, whereas the changes in Width parameter induce smoother variations within the generally achieved levels of performance. From the accuracy diagrams we can determine that in order to maintain levels of accuracy under 1° the Length parameter needs to be larger than 2.5 mm for the horizontal detection areas ($PS_1/PS_2$, see Fig. 8) and larger than 1 mm for vertical detection areas ($PS_3/PS_4$). In Table I, we show the optimum values that can be achieved for accuracy, with horizontal accuracy being slightly better than vertical. The calculated standard deviation does not allow for the observation of a clear effect, however, as we will see this trend (better horizontal accuracy than vertical) is observed in most other designs too, supporting the generality of this effect. For the case of crosstalk, we can observe that the Width parameter affects performance at a larger extent. From the respective diagrams we can determine that in order to push the horizontal-output crosstalk at levels lower than 10% we should select Length larger than 2 mm and Width larger than 5 mm. For the vertical-output crosstalk, we need Length larger than 1.5 mm and smaller than 9.5 mm, whereas for Width it is preferable to avoid the range [5 mm, 8 mm]. In Table II, we show the optimum achieved values for crosstalk. In this case, the optimum value for vertical-output crosstalk seems to be slightly better than horizontal, however, the opposite phenomenon seems to be the rule for the other three designs. The joint examination of Tables I-II provides very interesting information regarding the competing conditions that lead to optimization of horizontal accuracy (smaller Width, larger Length) and horizontal-output crosstalk (larger Width, smaller Length). Given that the horizontal accuracy and horizontal-output crosstalk are calculated for the same sensors ($PS_1/PS_2$), during a practical implementation, the final values for the parameters need to be chosen based on a suitable trade-off.

The same holds for vertical accuracy and vertical-output crosstalk, given that these measures are calculated for sensors $PS_3/PS_4$.

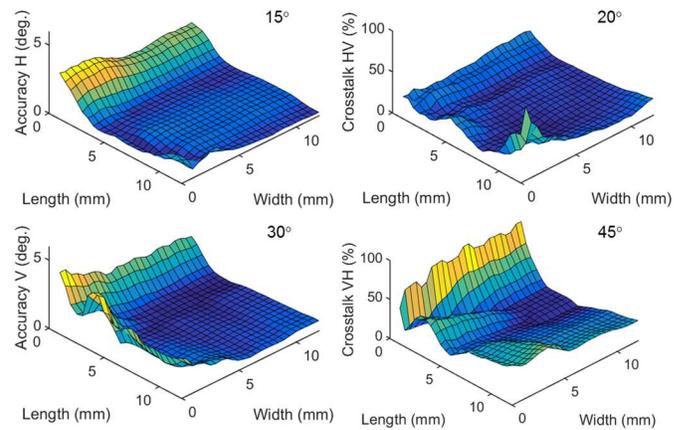

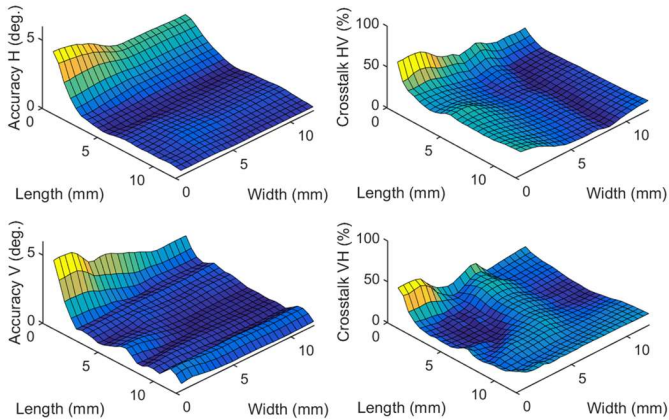

Fig. 9. Variation of accuracy and crosstalk as function of Length and Width parameters of the detection areas of design $D_1$.

In Fig. 10 we present the respective surface-diagrams for design $D_2$. For this design we vary the Length (L) and Width (W) parameters of the rectangle detection areas, and also, the Angle (A) that they form with the horizontal direction. Given that the optimum characteristics for accuracy and crosstalk can be achieved at different angles, each of the diagrams presents the Length-Width surface where the optimum was achieved and denotes the respective angle. An important characteristic of design $D_2$ is that the same detection areas ($PS_1/PS_2$, Fig. 8) are used both for the calculation of horizontal output (addition) and for the calculation of vertical output (subtraction). This makes the investigation for the optimum parameters even more challenging. From the current diagrams, we can determine that in order to reach accuracy levels under 1° in both directions (horizontal and vertical) the Length needs to be larger than 2.5 mm and the Width larger than 1 mm. The optimum values are achieved at angles of 15° and 30° for horizontal and vertical accuracy respectively. As we can observe in Table I, there are competing conditions that lead to optimization of horizontal and vertical accuracy. For the case of horizontal accuracy, the optimum performance is achieved for much larger Length than Width, whereas for achieving optimum vertical accuracy the detection areas have much larger Width than Length. It should be noticed that, practically, the much larger Width means that the inverted 'V' shape formed by the original detection areas (Fig. 8) becomes a normal 'V' shape, as shown in Fig. 13. For the case of crosstalk, we can see similar+ competing conditions. In specific, larger Length and smaller Width favors horizontal-output crosstalk and vice versa for the vertical-output crosstalk. Due to these competing conditions, in order to keep the crosstalk in levels close or under of 10% simultaneously for both outputs (horizontal and vertical) we need to limit the Length of the detection areas approximately within the range [3 mm, 7 mm] with the Width being larger than 6.5 mm.

Fig. 10. Variation of accuracy and crosstalk as function of Length, Width and Angle parameters of the detection areas of design $D_2$.

In the diagrams for design $D_3$, shown in Fig. 11, the behavior of accuracy and crosstalk is represented using curves (instead of surfaces) since in this case the only changing parameter is the Diameter (D) of the circularly modulated detection areas ($PS_1/PS_2$ for horizontal output, and $PS_3/PS_4$ for vertical output, see Fig. 8). For horizontal accuracy, we can see that levels under 1° can be achieved for Diameter larger than 3 mm, and levels of about 0.5° seem to be kept up to the upper bound of the tested range. Vertical accuracy reaches levels less than 1° for Diameter larger than 2.5 mm but then degrades again to levels over 1° for Diameter larger than 10.5 mm. The optimum values for horizontal and vertical accuracy are shown in Table I, and we can observe that they are slightly worse than the previous designs, with the horizontal output being once again more accurate than the vertical. For the case of horizontal-output crosstalk it is possible to achieve levels under 10% when using Diameter larger than 5.5 mm. The behavior of vertical-output crosstalk is inferior, specifically, it can be observed that the vertical-output crosstalk does not reach levels under 10%, with the optimum value being 14.5%. This optimum value is achieved for Diameter of 5.5 mm. Also, from Fig. 11 we can see that the Diameter needs to be larger than 4 mm and smaller than 8.5 mm to keep vertical-output crosstalk lower than 20%.

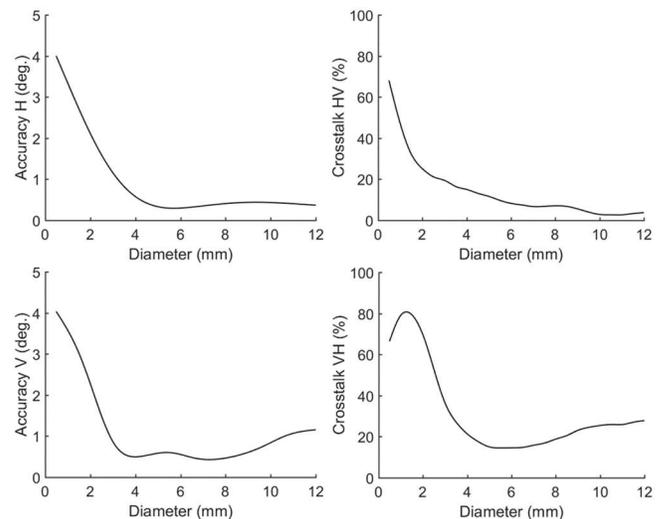

Fig. 11. Variation of accuracy and crosstalk as function of Diameter parameter of the detection areas of design $D_3$.

For the case of design $D_4$, the parameters that we change are the Diameter (D) and vertical Position (Pos. Y) of the detection array of the horizontal output, and the Diameter and horizontal Position (Pos. X) of the detection array of the vertical output. In Fig. 12, we can observe the behavior of accuracy and crosstalk when we change the respective parameters. We can see that the positioning of the horizontal detection areas slightly upwards (+) is preferable for achieving optimum horizontal accuracy. Vertical accuracy seems to be more symmetric across the positioning axis but it becomes optimum for positioning of the vertical array closer to the nasal area (−). For the case of crosstalk, the positioning of the horizontal detection areas at the upper part of the eye once again leads to optimized performance. The Diameter needs to be larger than 3 mm and positioning needs to be slightly upwards to lead to horizontal-output crosstalk lower than 10%. The vertical-output crosstalk is more symmetric with regard to positioning, and levels close to 11% are marginally reached for Diameter larger than 11.5 mm and positioning close to the pupil center. The optimum values and the respective parameters for Design D4 can be overviewed in Table II.

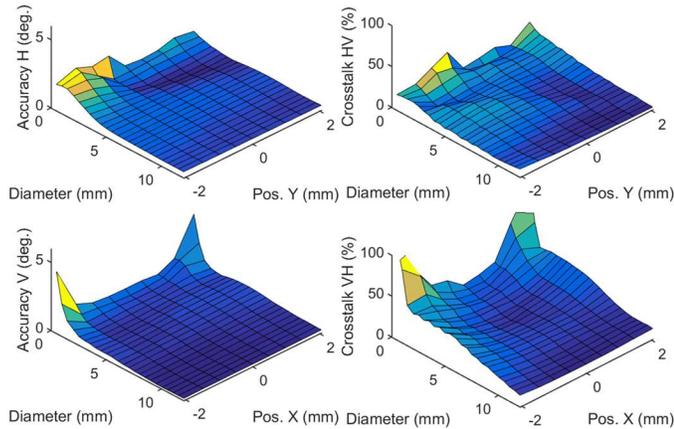

Fig. 12. Variation in accuracy and crosstalk as function of Diameter and Position parameters of the detection areas of design $D_4$.

TABLE I
Optimum values for accuracy and the corresponding design parameters.

|  | $D_1$ | $D_2$ | $D_3$ | $D_4$ |
|---|---|---|---|---|
| **Acc. H (°)** | 0.12 (0.07) | 0.09 (0.06) | 0.31 (0.17) | 0.22 (0.11) |
| design parameters | W: 3.0 mm<br>L: 5.5 mm | W: 1.5 mm<br>L: 9.0 mm<br>A: 15° | D:5.0 mm | D: 3.0 mm<br>$P_y$:1.0mm |
| **Acc. V (°)** | 0.17 (0.13) | 0.13 (0.09) | 0.42 (0.41) | 0.13 (0.12) |
| design parameters | W: 9.5 mm<br>L: 9.5 mm | W:11.0mm<br>L: 4.0 mm<br>A: 30° | D: 7.0 mm | D: 4.9 mm<br>$P_x$:-2 mm |

TABLE II
Optimum values for crosstalk and the corresponding design parameters.

|  | $D_1$ | $D_2$ | $D_3$ | $D_4$ |
|---|---|---|---|---|
| **Cross. HV (%)** | 2.8 (2.5) | 1.5 (1.4) | 2.6 (2.0) | 3.1 (2.6) |
| design parameters | W: 9.0 mm<br>L: 3.0 mm | W: 5.5 mm<br>L: 8.0 mm<br>A: 20° | D:11.0mm | D:11.0mm<br>$P_y$:1.5mm |
| **Cross. VH (%)** | 1.4 (1.5) | 1.9 (2.4) | 14.5 (7.6) | 11.1 (6.8) |
| design parameters | W: 3.5 mm<br>L: 5.0 mm | W:11.5mm<br>L: 6.0 mm<br>A: 45° | D: 5.5 mm | D:12.0mm<br>$P_x$: 0.5 mm |

Our investigation clearly shows that during the practical selection of the parameters of a PSOG design we will need to counterbalance the competing requirements that lead to optimization the accuracy and crosstalk measures. For this reason, in addition to the results showing the parameter values that optimize each measure separately (Tables I-II), we performed a simple multi-objective optimization procedure to find a set of parameters that can provide an overall satisfactory level of performance both for accuracy and crosstalk (trade-off parameters). The performed multi-objective optimization procedure, finds the points in parameter space (surface-diagrams or curve-diagrams) where the minimum values of the different measures occur, and then, iteratively expands these areas (or lines) until a meeting point is found. The important point in this procedure is that rate of expansion for the different measures is always kept to be inversely proportional to the rate of relative increase of their values (e.g. 1%, 2% etc.). Designs $D_1$, $D_3$ and $D_4$ use different sets of sensors for horizontal and vertical outputs, so, the multi-objective optimization is performed separately for measure-pairs: **Acc. H/Cross. HV** and **Acc. V/Cross VH**. For design $D_2$, on the other hand, the optimization needs to be done simultaneously for all measures. In Table III, we present the estimated trade-off parameters for each design along with the respective values for accuracy and crosstalk, and in Fig. 13 we provide a graphical view of the transformed detection areas when using the trade-off parameters. We can observe that the increase in values of accuracy and crosstalk (due to trade-off) stays at relatively reasonable levels. Also, the comparative inspection of the parameter values from Tables I, II, and III, reveals the factors that dictate the selection of trade-off parameters. For example, for design $D_1$ and for measure-pair **Acc. H/Cross. HV** the Width trade-off parameter is dictated by **Cross. HV** (Table II) whereas the Length trade-off parameter is dictated by **Acc. H** (Table I). For measure-pair **Acc. V/Cross. VH** the Length trade-off parameter is somewhere between the values from Tables I-II, whereas the Width trade-off parameter is larger than both values from Tables I-II. This is due to the fact that the region of low **Acc. V** values drives the estimation of trade-off Width parameter into the second region (valley) of low **Cross. VH** values (see Fig. 9). Analogous comparative observations of the trade-off values along with the inspection of respective figures can explain the parameter selection for the other PSOG designs.

TABLE III
Trade-off values for accuracy and crosstalk and the corresponding design parameters.

|  | $D_1$ | $D_2$ | $D_3$ | $D_4$ |
|---|---|---|---|---|
| **Acc. H (°)** | 0.17 (0.12) | 0.28 (0.09) | 0.38 (0.12) | 0.33 (0.16) |
| **Cross. HV (%)** | 3.5 (2.7) | 4.8 (3.5) | 3.3 (2.0) | 4.7 (2.6) |
| trade-off parameters | W: 9.0 mm<br>L: 5.0 mm | W:11.0 mm<br>L: 5.0 mm<br>A: 15° | D:11.5 mm | D:10.5 mm<br>$P_y$:2.0mm |
| **Acc. V (°)** | 0.25 (0.17) | 0.40 (0.23) | 0.42 (0.41) | 0.16 (0.12) |
| **Cross. VH (%)** | 2.4 (2.8) | 4.3 (4.5) | 14.8 (6.8) | 14.0 (11.2) |
| trade-off parameters | W:11.0 mm<br>L: 6.5 mm | W:11.0 mm<br>L: 5.0 mm<br>A: 15° | D: 6.5 mm | D:11.5 mm<br>$P_x$:-1.5mm |

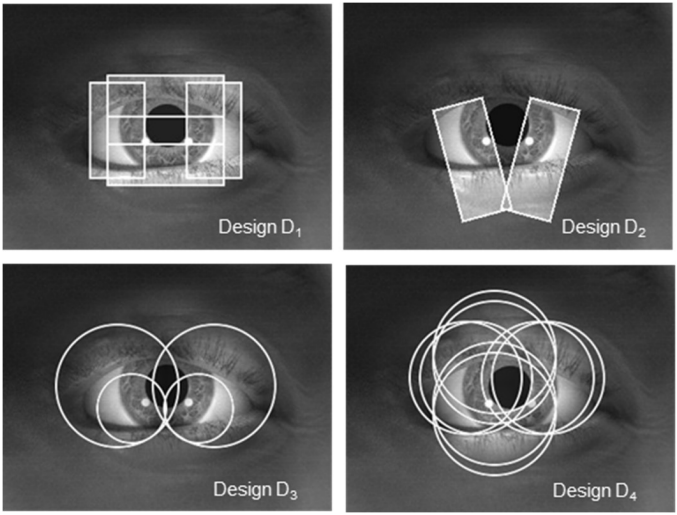

Fig. 13. The resulting designs when using the estimated trade-off parameters.

### B. The Effects of Sensor Shifts on Eye-tracking Performance

When the sensor is shifted away from the initial position where the calibration was done, then, the areas tracked by the sensors slightly change. This means that for the same eye positions the raw output values of the PSOG design will be changed, and the mapping performed by calibration function will result in accuracy errors. In this section, we investigate the behavior of PSOG designs when such sensor shifts occur after the calibration process, and we demonstrate the extent of the accuracy error when increasing the magnitude of sensor shifts.

The experiments for sensor shifts are performed using the simulation framework described in Section III. To simulate sensor shifts, we move the camera model of the 3-D scene at a number of predefined positions. The selected range of sensor shifts is [-2 mm, 2 mm] (horizontal and vertical) with a step of 0.5 mm. In order to perform a dense scan of eye positions, in this case we use a set of predefined ground truth eye positions to rotate the eye model. The selected range of eye positions is [-10°, 10°] (horizontal and vertical) with step of 0.5°. Please notice that during the presentation of results we use the following sign conventions: horizontal eye movements are positive when the (left) eye moves towards the nasal area, and negative when it moves away from the nasal area. Vertical eye movements are positive when the eye moves downwards, and negative when it moves upwards. Horizontal sensor shifts are positive when the sensor moves away from the nasal area, and negative when the sensor moves towards the nasal area. Vertical sensor shifts are positive when the sensor moves upwards, and negative when the sensor moves downwards.

The sensor shift experiments are performed for designs $D_{1-4}$ using the trade-off parameters determined in previous section. We present the results from the simulations in the form of curve-cluster diagrams, shown in Fig. 14-17, where for each PSOG design we demonstrate the estimated eye positions when sensor shifts occur as function of ground truth eye positions. In each figure we show four diagrams covering different combinations of sensor shifts and eye movements, specifically, horizontal eye movements when horizontal sensor shifts occur (denoted as H-H), vertical eye movements when horizontal sensor shifts occur (denoted as V-H), horizontal eye movements when vertical sensor shifts occur (denoted as H-V), vertical eye movements when vertical sensor shifts occur (denoted as V-V).

We additionally present a summarizing measure of the observed behavior in each curve-cluster. We calculate the Mean Absolute Error/MAE (across eye movement range) between each curve corresponding to a specific sensor shift $i$ and the original curve (zero sensor shift), as shown in Eq. (10). In Fig. 18, we present the calculated values for MAE-between-curves for all designs.

$$MAE_i^{between-curve} = \sum_{j=1}^{M}\left|y_j^{SnMv_i} - y_j^{SnMv_0}\right|/M \qquad (10)$$

Where $M$ is the number of samples in the curves and $y_j$ the corresponding value for sample $j$.

In Fig. 14 we present the curve-cluster diagrams for design $D_1$. We can see that the sensor shifts result in a translation of the curves (loss of accuracy) and that the effect is more prominent when sensor shifts occur at the same direction with the eye movement (combinations H-H and V-V). We can see that the linearity of the curves gradually breaks, and for the case of combination V-V we can see a strong break in linearity for downward sensor shifts of 2 mm when upward eye movements occur. This phenomenon can be attributed to the fact that the lower detection area ($PS_4$) in this case points at the lower eyelid region, which has relatively low mobility and reflects more light for upward eye movements, something that is not the case of the upper eyelid. The combinations V-H and H-V are less affected from sensor shifts, and specifically, we can observe that the curves for combination V-H overlap at a large extent, which reveals the robustness of the design to horizontal sensor shifts when vertical eye movements occur. In Fig. 18 (top-left) we can observe the MAE-between-curves for all combinations. For H-H and V-V, the error goes over 3° for sensor shifts with magnitude larger than 1 mm. The better eye-tracking behavior for combinations V-H and H-V can be verified by the respective MAE-between-curves values that remain under 2° for all tested range of sensor shifts.

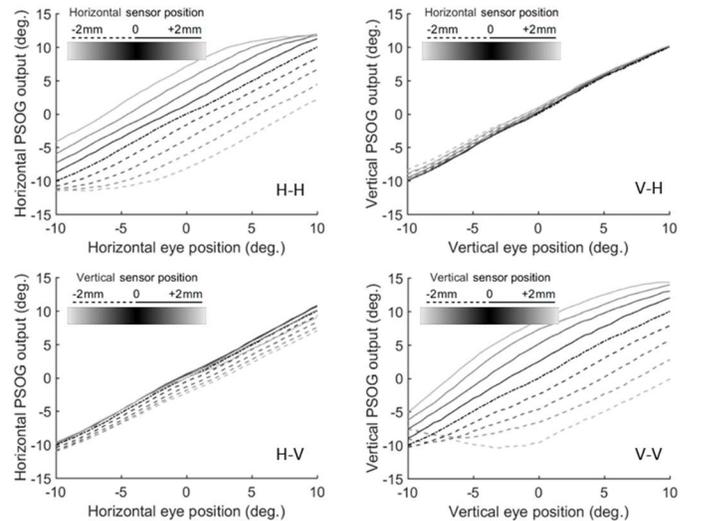

Fig. 14. Curve-clusters of eye-tracking accuracy when sensor shifts occur, for the case of design $D_1$.

In Fig. 15 we can observe the corresponding curve-clusters for design $D_2$. As previously, the H-H and V-V combinations seem to be affected more heavily, and the loss of accuracy is much larger for combination V-V (the curves are further apart). It should be emphasized, though, that despite the steeper degradation of accuracy for these combinations (over 5° for sensor shifts larger than 1 mm, as shown in Fig. 18, top-right) the linearity of the V-V curves seems to be preserved better than the previous design, and this can make the modeling of the curve-cluster easier. Such modeling can be used in a scenario where an external mechanism is employed to estimate sensor shift and a composite calibration function is used to correct the output. For the V-H and H-V combinations the accuracy loss is milder, however, the behavior seems to be less robust when compared to the previous design (especially for V-H curves). As we observe in Fig. 18, for V-H combination the MAE-between-curves can be larger than 2° for sensor shifts larger than 1mm. A possible reason for this sensitivity is the operation of 'addition' used to calculate the vertical output in design $D_2$. Another interesting observation for this design is that for H-V combination the sensor shifts result in a rotation transformation of the curves instead of parallel translation.

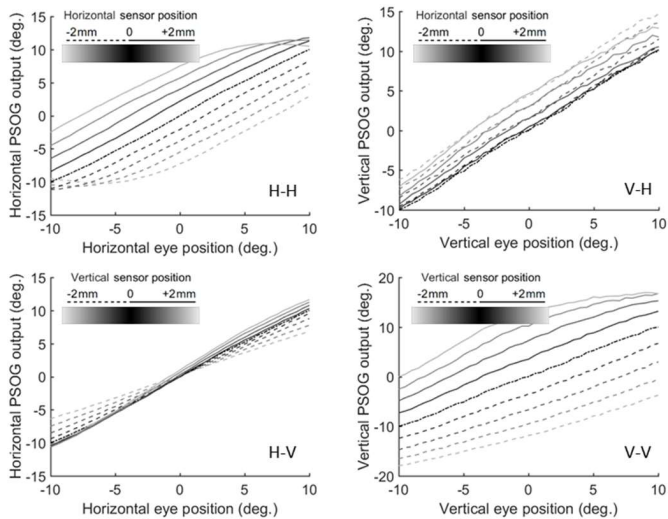

Fig. 15. Curve-clusters of eye-tracking accuracy when sensor shifts occur, for the case of design $D_2$.

In Fig. 16 we show the curve-cluster diagrams for design $D_3$, and in Fig. 18 (bottom-left) we can observe the respective values for MAE-between-curves. The large accuracy loss for H-H and V-V combinations results in MAE-between-curves that can be over 4° for sensor shifts of magnitude over 1 mm, and reaches 6°-8° for sensor shifts of 2 mm. Although the MAE-between-curves for H-H combination has slightly steeper slope than V-V (reaches larger values sooner), the diagrams of Fig. 16 show that the eye-tracking curves present smoother and more symmetric behavior. In contrast, the curves for V-V combination successively lose their linearity (and reverse their slope) for upward sensor shifts when downward eye movements occur. The V-H and H-V combinations present less accuracy loss. For the V-H combination the linearity breaks for sensor shifts towards the nasal area and large downward eye movements. A possible cause of this phenomenon is that the downward eyelid movement generates larger reflection and the reason why this effect is observed for sensor shifts towards the nasal area can be attributed to the generally larger amount of skin and eyelid reflection at the nasal area when compared to the temporal area.

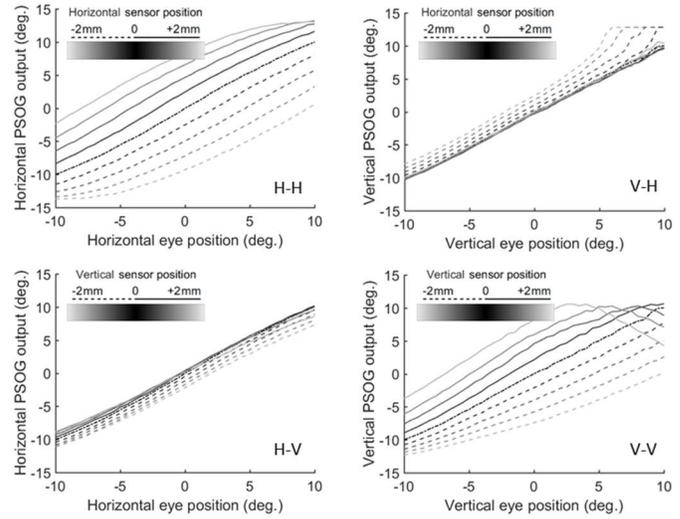

Fig. 16. Curve-clusters of eye-tracking accuracy when sensor shifts occur, for the case of design $D_3$.

In Fig. 17 we show the curve-clusters for design $D_4$. The accuracy loss for combinations H-H and V-V seems to be at similar levels with the other designs, with larger loss for the V-V combination. A positive aspect of this design, though, is that for both of these combinations the curves are translated homogeneously and the linearity is preserved at a large extent. As we described earlier, this implies an easier modeling of the behavior of the output during sensor shifts, a feature that can be valuable for the alternative scenario of external sensor shift estimation and correction. Furthermore, the specific design appears to have the best behavior for the V-H and H-V combinations, as portrayed by the small dispersion of the curves. The respective MAE-between-curves shown in Fig. 18 (bottom-right) further demonstrate this behavior, with accuracy loss of well under 2° for the whole range of sensor movements.

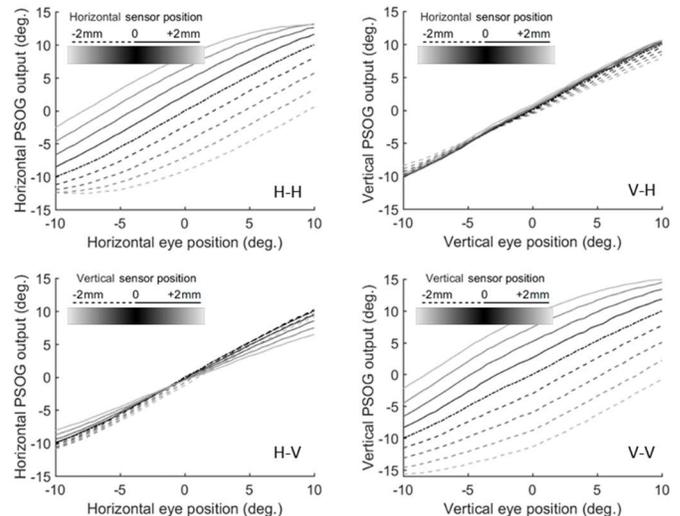

Fig. 17. Curve-clusters of eye-tracking accuracy when sensor shifts occur, for the case of design $D_4$.

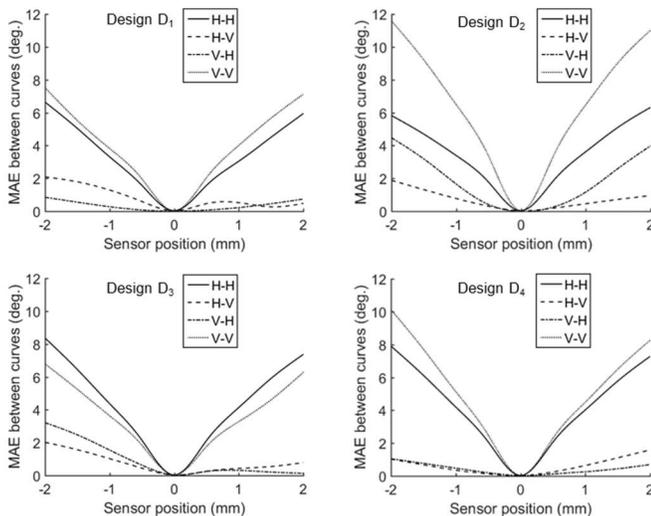

Fig. 18. Mean Absolute Error (MAE) between curves for different magnitudes of sensor shift.

*C. Limitations and Further Extensions*

The current experimentation aimed to overview the properties of different PSOG designs under a common basis and to systematically investigate the variation of their eye-tracking characteristics when changing the design parameters. The developed simulation framework served very well this purpose and the next step in this line of research should be to address some of the current limitations. First, the extension of the experiments by using eye images from a large number of subjects can strengthen the generality of the current observations, and additionally, it would allow to explore the effects on eye-tracking quality from dynamic phenomena like blinks, squinting, and occlusion by eyelashes. Second, our current experiments examined combinations of pure horizontal and vertical eye movements and sensor shifts. Although these types of movements are very basic for the understanding of the properties of a design, a next and more challenging step would be to examine the scenario of oblique eye movements and sensor shifts, and of rotational sensor shifts. Last, the used range of eye movements was [-10°, 10°] and the used range of sensor shifts was [-2 mm, 2 mm]. These ranges were chosen to facilitate the homogeneous examination of the properties of all tested PSOG designs, allowing thus for more interpretable results. A further extension of this work can pursue the limits of traceable eye movement ranges, and examine alternative approaches for the improvement of the eye-tracking properties during sensor shifts, e.g. addition of more sensors, different ways to combine the sensor outputs etc.

## V. Conclusion

In this work we examined and assessed the characteristics of different photosensor oculography (PSOG) designs. For this purpose, we used a simulation framework that allowed us to iteratively change the parameters of the eye detection areas of PSOG designs, and to demonstrate the resulting effects on accuracy (mean absolute error compared to ground truth) and crosstalk (interference in horizontal and vertical output). Our experimentation showed that in most cases the simultaneous optimization of these two measures of eye-tracking quality needs to be a balancing act aiming to find a satisfactory trade-off among the competing requirements for the design parameters. This finding outlines the importance of the current research, since the presented clues regarding the behavior when changing the PSOG design parameters towards one or the other direction can provide valuable feedback during the creation of a PSOG system. Such clues can be useful for debugging issues due to bad selection of design parameters, and for steering the engineering of a PSOG system towards the direction of desired performance levels. We additionally demonstrated the effects on performance when sensor shifts occur. The results showed that even though all PSOG designs are strongly affected by sensor shifts there are differences in the rate of accuracy degradation, in the behavior for different combinations of eye movements and sensor shifts, and in the easiness of modeling the resulting behavior. All these factors should be weighted properly when trying to improve the sensor-shift robustness using different approaches (e.g. cover larger area, use of external sources etc.). In overall, we believe that the current findings can be a valuable contribution for triggering the research on photosensor oculography even further, and lead to improvements that would allow to deploy the full potential of this technology.